\documentclass[english]{llncs}
\usepackage{ae,aecompl}
\usepackage[T1]{fontenc}
\usepackage[latin9]{inputenc}
\usepackage{float}
\usepackage{amstext}
\usepackage{amssymb}
\usepackage{graphicx}

\makeatletter

\providecommand{\tabularnewline}{\\}
\floatstyle{ruled}
\newfloat{algorithm}{tbp}{loa}
\providecommand{\algorithmname}{Algorithm}
\floatname{algorithm}{\protect\algorithmname}

\@ifundefined{showcaptionsetup}{}{%
 \PassOptionsToPackage{caption=false}{subfig}}
\usepackage{subfig}
\makeatother

\usepackage{babel}
\begin{document}

\title{Quantum Latent Semantic Analysis \thanks{This paper was originally presented at the  2011 ACM SIGIR International Conference on the Theory of Information Retrieval ICTIR2011 }}

\author{Fabio A. Gonz\'{a}lez and Juan C. Caicedo \\
 \{ fagonzalezo, jccaicedoru\}@unal.edu.co\\
}

\institute{Mindlab Research Group\\
Computing Systems and Industrial Engineering Dept.\\
National University of Colombia}
\maketitle
\begin{abstract}
The main goal of this paper is to explore latent topic analysis (LTA),
in the context of quantum information retrieval. LTA is a valuable
technique for document analysis and representation, which has been
extensively used in information retrieval and machine learning. Different
LTA techniques have been proposed, some based on geometrical modeling
(such as latent semantic analysis, LSA) and others based on a strong
statistical foundation. However, these two different approaches are
not usually mixed. Quantum information retrieval has the remarkable
virtue of combining both geometry and probability in a common principled
framework. We built on this quantum framework to propose a new LTA
method, which has a clear geometrical motivation but also supports
a well-founded probabilistic interpretation. An initial exploratory
experimentation was performed on three standard data sets. The results
show that the proposed method outperforms LSA on two of the three
datasets. These results suggests that the quantum-motivated representation
is an alternative for geometrical latent topic modeling worthy of
further exploration.

\keywords{Quantum mechanics \and Quantum information retrieval \and Latent
semantic analysis \and  Latent semantic indexing \and Latent topic analysis \and
Singular value decomposition \and Probabilistic latent semantic analysis}

\end{abstract}

\section{Introduction}

Since its inception, latent topic analysis\footnote{For the remaining part of the text we will use the term \emph{latent
topic analysis} to allude the general modeling strategy avoiding confusion
with \emph{latent semantic analysis} which refers to the particular
method.} (LTA) (also known as latent semantic analysis/indexing) has been
a valuable technique for document analysis and representation in both
information retrieval (IR) and machine learning. The main assumption
behind LTA is that the observed term-document association may be explained
by an underlying latent topic structure. Different methods for latent
topic analysis have been proposed, the most prominent include: latent
semantic analysis (LSA) \cite{Deerwester1990}, probabilistic latent
semantic analysis (PLSA) \cite{Hofmann1999}, and latent Dirichlet
allocation (LDA) \cite{Blei2003}. LSA was the first latent analysis
method proposed and its approach is geometrical in nature, while PLSA
and LDA have a sound probabilistic foundation. 

Quantum information retrieval (QIR) \cite{VanRijsbergen2004,Song2010},
is a relatively new research area that attempts to provide a foundation
for information retrieval building on the mathematical framework that
supports the formulation of quantum mechanics (QM). QIR assimilates
the traditional vector space representation to Hilbert spaces, the
fundamental concept in QM. Notions such as system state, measurement,
uncertainty and superposition are interpreted in the context of IR.
QIR is been actively researched and some results suggest that it can
go beyond an interesting analogy to become a valuable theoretical
and methodological framework for IR \cite{Song2010}.

The main goal of this paper is to explore latent topic analysis in
the context of QIR. Same as in the vector space model, QIR represents
documents/queries as vectors in a vector space (more precisely, a
Hilbert space), however, QIR exploits the subspace structure of the
Hilbert space and corresponding probability measures to define important
IR notions, such as relevance, in a principled way \cite{VanRijsbergen2004}.
A question that emerges is whether the richer QIR representation could
provide new insights into the latent topic analysis problem. One important
motivation for this question is the fact that QIR naturally combines
both geometry and probability. Latent topic analysis methods proposed
so far are either geometrical or probabilistic in nature, but not
both. A quantum-motivated latent semantic analysis method could potentially
combine both perspectives.

Some works in QIR \cite{DiBuccio,Piwowarski2010,Widdows2009,Melucci2008}
have already suggested the relationship between LTA and a quantum-based
representation of documents. Up to our knowledge, there has not been
proposed yet an original LTA algorithm in a quantum representation
context. The work of Melucci \cite{Melucci2008} probably is the closest
one to the work presented in this paper. In that work, a framework
for modeling contexts in information retrieval is presented. The framework
uses both a quantum representation of documents and LSA to model latent
contexts, but do not propose a new LTA method. 

This paper proposes a new LTA method, quantum latent semantic analysis
(QLSA). The method starts from a quantum-motivated representation
of a document set in a Hilbert space $H$. The latent topic space
is modeled as a sub-space of $H$, where the document set is projected.
The method is analyzed from geometrical and probabilistic points of
view, and compared with LSA and PLSA. An exploratory experimentation
was performed to evaluate how the quantum-motivated representation
impacts the performance of the method. The results show that the method
outperforms LSA on two of the three datasets, and we hypothesize that
it is due to an improved quantum representation. 

The paper is organized as follows: Section 2 provides a brief overview
of quantum information retrieval; Section 3 describes the method and
discusses its similarities and differences with LSA and PLSA; Section
3 covers the exploratory experimental evaluation of the method; finally,
Section 4 presents some conclusions and the future work.

\section{Quantum Information Retrieval}

QIR provides an alternative foundation for information retrieval.
The main ideas were initially proposed by Van Rijsbergen \cite{VanRijsbergen2004},
and different subsequent works have contributed to the continuous
development of the area. The main idea in QIR is to use the quantum
mechanics formalism to deal with fundamental information retrieval
concepts exploiting clear analogies between both areas. For instance,
a quantum system state is represented by a wave function, which can
be seem as a finite or infinite complex vector indexed by a continuous
or discrete variable (usually representing space or momentum). In
a vector space model, documents are represented by vectors, but in
this case they are finite real vectors indexed by a discrete variable
that represents text terms. In the next paragraphs we will briefly
present some basic concepts from QIR that are necessary to introduce
the proposed method.

Lets $D=\{d_{i}\}_{i=1\dots n}$ be a set of documents, $T=\{t_{j}\}_{j=1\dots m}$
be a set of terms, and $TD=\{td_{ji}\}$ be the corresponding term-document
matrix. The quantum representation of a document $d_{i}$ is given
by a wave function $\varphi_{i}$ defined by:
\[
\varphi_{i}(j)=\sqrt{\frac{td_{ji}}{\sum_{j=1}^{m}td_{ji}}}\textrm{, for all }j=1\dots m,
\]
This representation has the following convenient properties:

\[
\forall i,\left\Vert \varphi_{i}\right\Vert =1
\]

\begin{equation}
<\varphi_{i},\tau_{j}>^{2}=P(t_{j}|d_{i})\label{eq:term-doc-prob}
\end{equation}

where $<\cdot,\cdot>$ is the dot product operator, $\tau_{j}$ is
the wave function of the term $t_{j}$ corresponding to a unitary
vector with a one in the $j$-th position. This representation corresponds
in fact to a representation of the documents in the term space, which
we will call $H$ and whose basis is $\{\left|\tau_{j}\right\rangle \}_{j=1\dots m}$. 

Dirac notation is a convenient notation formalism extensively used
in quantum mechanics. The two basic building blocks of Dirac notation
are the \emph{bra} and the \emph{ket}, notated respectively as $\left\langle \varphi\right|$
and $\left|\beta\right\rangle $. A ket represents a vector in a Hilbert
space and a bra a function from the Hilbert space to a real (or complex)
space. The application of a bra to a ket coincides with the dot product
of the corresponding vectors and is notated $\left\langle \varphi|\beta\right\rangle $.
In a finite-dimensional Hilbert space, a bra may be seem as a row
vector and a ket as a column vector, in this case the application
of a bra to a ket would correspond to a conventional matrix multiplication. 

A bra and a ket can be composed in a reverse way, $\left|\beta\right\rangle \left\langle \varphi\right|$,
and this can be interpreted as the outer product of the corresponding
vectors. This is useful, for instance, to define notions such as subspace
projectors. A subspace is determined by a basis that generates it
or by a projector operator that projects any vector in the space to
the subspace. If the basis of a given subspace $S$ is $\{\beta_{1},\dots,\beta_{m}\}$,
the corresponding projector is $P_{s}=\sum_{i=1\dots m}\left|\beta_{i}\right\rangle \left\langle \beta_{i}\right|$.
Projectors with trace one are called density operators and have an
important role in quantum mechanics, they are used to represent the
statistical state of a quantum system.

Using Dirac notation the second property in Eq. \ref{eq:term-doc-prob}
can be expressed as $\left\langle \varphi_{i}|\tau_{j}\right\rangle {}^{2}=P(t_{j}|d_{i})$.
This property can be interpreted, in a QIR context, as the density
operator $\rho_{i}=\left|\varphi_{i}\right\rangle \left\langle \varphi_{i}\right|$
(corresponding to the document $d_{i}$) acting on the subspace $P_{\tau_{j}}=\left|\tau_{j}\right\rangle \left\langle \tau_{j}\right|$
(which is induced by the term $t_{j}$) according to the rule:
\[
P(P_{\tau_{j}}|\rho_{i})=\textrm{tr}(\rho_{i}P_{\tau_{j}})=\textrm{tr}(\left|\varphi_{i}\right\rangle \left\langle \varphi_{i}|\tau_{j}\right\rangle \left\langle \tau_{j}\right|)=\left\langle \varphi_{i}|\tau_{j}\right\rangle {}^{2},
\]
 where $tr(\cdot)$ is the matrix trace operator. The above procedure
could be extended to more complex subspaces, i.e., with dimension
higher than one.

\section{Quantum Latent Semantic Analysis}

In general, LTA modeling assumes that the high diversity of terms
in a set of documents may be explained by the presence or absence
of latent semantic topics in each document. This induces a new document
representation where documents are projected to a latent topic space
by calculating the relative degree of presence of each topic in each
document. Since the set of latent semantic topics is usually one or
two orders of magnitude smaller than the set of terms, the effective
dimension of the latent topic space is smaller than the dimension
of the original space, and the projection of the document to it is,
in fact, a dimensionality reduction process. 

A latent topic space is a subspace $S$ of $H$ defined implicitly
by its projector as:
\[
P_{S}=\sum_{k=1}^{r}\left|\sigma_{k}\right\rangle \left\langle \sigma_{k}\right|,
\]
 where $\{\left|\sigma_{k}\right\rangle \}_{k=1\dots r}$ is an orthonormal
basis of the sub-space $S$ and each $\left|\sigma_{k}\right\rangle $
corresponds to the wave function of a latent topic $z_{k}$. A projection
of a document represented by $\left|\varphi_{i}\right\rangle $ on
the latent space is given by:
\[
\left|\bar{\varphi_{i}}\right\rangle =P_{S}\left|\varphi_{i}\right\rangle .
\]

From a quantum mechanics perspective, this projection can be interpreted
as the measurement of the observable corresponding to $S$ on the
system state $\left|\varphi_{i}\right\rangle $. This measurement
will make the state of the system collapse to a new state $\left|\hat{\varphi_{i}}\right\rangle =\frac{\left|\bar{\varphi_{i}}\right\rangle }{\left\Vert \left|\bar{\varphi_{i}}\right\rangle \right\Vert }$.
Accordingly, the conditional probability of latent topic $z_{k}$
given a document $d_{i}$ represented in the latent space can be calculated
by:
\[
P(z_{k}|d_{i})=\left\langle \hat{\varphi_{i}}|\sigma_{k}\right\rangle ^{2}=\frac{\left\langle \varphi_{i}|\sigma_{k}\right\rangle ^{2}}{\left\Vert P_{S}\left|\varphi_{i}\right\rangle \right\Vert ^{2}}.
\]

Now, the main problem is to find an appropriate latent semantic topic
space $S$. This can be accomplished by imposing some conditions.
In particular, we expect that the latent topic representation loses
as few information as possible and be as compact as possible. This
can be expressed trough the following optimization problem:
\[
\min_{{S\atop dim(S)=r}}\sum_{i=1}^{n}\left\Vert \left|\bar{\varphi_{i}}\right\rangle -\left|\varphi_{i}\right\rangle \right\Vert ^{2}=\min_{{S\atop dim(S)=r}}\sum_{i=1}^{n}\left\Vert P_{S}\left|\varphi_{i}\right\rangle -\left|\varphi_{i}\right\rangle \right\Vert ^{2}
\]

This problem is solved by performing a singular value decomposition
(SVD) on the matrix formed by the vectors corresponding to the wave
functions of the documents in the document set. Specifically, a matrix
where the $i$-th column corresponds to the ket $\left|\varphi_{i}\right\rangle $,
$\Phi=\left[\varphi_{1}\dots\varphi_{n}\right]$, with 
\[
\Phi=U\Sigma V^{T},
\]
its SVD decomposition. The columns of $U=\left[\sigma_{1}\dots\sigma_{r}\right]$,
correspond to the vectors of an orthonormal basis of the latent subspace
$S$. The process is summarized in Algorithm 1.

\begin{algorithm}[t]
Quantum-LSA($TD$,$r$)

~~~$TD=\{td_{ij}\}$: term-document matrix with $i=1\dots m$ and
$j=1\dots n$.

~~~$r$: latent topic space dimension\medskip{}

1: Build the document wave function matrix $\Phi\in\mathbb{R}^{m\times n}$
setting 
\[
\Phi_{ij}=\sqrt{\frac{td_{ji}}{\sum_{j=1}^{m}td_{ji}}}
\]

2: Perform a SVD of $\Phi=U\Sigma V^{T}$

3: Select the first $r$ columns of $U$, $\{\sigma_{1}\dots\sigma_{r}\}$,
corresponding to the $r$ principal Eigenvectors of $\Phi\Phi^{T}$.

4: Project each document wave function $\left|\varphi_{i}\right\rangle =\Phi_{\cdot i}$
\[
\left|\bar{\varphi_{i}}\right\rangle =\sum_{k=1}^{r}\left|\sigma_{k}\right\rangle \left\langle \sigma_{k}\right|\left|\varphi_{i}\right\rangle 
\]

5: Normalize the vector 
\[
\left|\bar{\varphi_{i}}\right\rangle =\frac{\left|\bar{\varphi_{i}}\right\rangle }{\left\Vert \left|\bar{\varphi_{i}}\right\rangle \right\Vert }
\]

6: The smoothed representation of a document $d_{i}$ in the term
space is given by 
\[
P(t_{j}|d_{i})=\bar{\varphi_{i}}(j)^{2}
\]

7: The document representation in the latent topic space is given
by
\[
P(z_{k}|d_{i})=\left\langle \bar{\varphi_{i}}|\sigma_{k}\right\rangle ^{2}
\]
 
\[
\]

\caption{Quantum latent semantic analysis}
\end{algorithm}

\subsection{QLSA vs LSA\label{subsec:QLSA-vs-LSA}}

Both QLSA and LSA use SVD as the fundamental method to find the latent
space. However, there is an important difference: LSA performs the
SVD decomposition of the original term-document matrix, whereas QLSA
decomposes the document wave function matrix, whose entries are proportional
to the square root of the original term-document matrix. This makes
QLSA a different method, since the decomposition is happening on a
different representation space. 

Both methods have a clear geometrical motivation, however QLSA has,
in addition, a natural probabilistic interpretation. LSA produces
a representation that may include negative values, this has been pointed
as a negative characteristic of latent topic representations based
on SVD \cite{Lee1999,xu03document}, since a document may be represented
by both the presence and the absence of terms or topics in it. QLSA,
in contrast, always produces positive values when documents are mapped
back to the term/topic space.

\subsection{QLSA vs PLSA}

The approach followed by PLSA is quite different to the one of QLSA.
PLSA has a strong statistical foundation that models documents as
a mixture of term probabilities conditioned on a latent random variable
\cite{Hofmann1999}. The parameters of the model are estimated by
a likelihood maximization process based on expectation maximization.
The mixture calculated by PLSA induces a factorization of the original
term-document matrix:
\begin{equation}
P(t_{j}|d_{i})=\sum_{k=1}^{r}P(t_{j}|z_{k})P(z_{k}|d_{i}),\label{eq:PLSA-decomposition}
\end{equation}
where $P(t_{j}|z_{k})$ codifies the latent topic vectors and $P(z_{k}|d_{i})$
corresponds to the representation of documents on the latent space.

QLSA also induces a factorization, but of the matrix formed by the
wave functions corresponding to the documents in the set. To illustrate
this lets check how the wave function of a document $d_{i}$ is codified
by QLSA:
\begin{eqnarray}
\varphi_{i}(j) & = & \left\langle \tau_{j}|\varphi_{i}\right\rangle \nonumber \\
 & \approx & \left\langle \tau_{j}|\hat{\varphi_{i}}\right\rangle \nonumber \\
 & = & \frac{\left\langle \tau_{j}\right|P_{S}\left|\varphi_{i}\right\rangle }{\left\Vert P_{S}\left|\varphi_{i}\right\rangle \right\Vert }\nonumber \\
 & = & \sum_{k=1}^{r}\left\langle \tau_{j}|\sigma_{k}\right\rangle \frac{\left\langle \sigma_{k}|\varphi_{i}\right\rangle }{\left\Vert P_{S}\left|\varphi_{i}\right\rangle \right\Vert }\label{eq:wave-function-cod-QLSA}
\end{eqnarray}

Eq. \ref{eq:wave-function-cod-QLSA} induces a factorization of the
document wave function matrix $\Phi$ into two matrices, one codifying
the latent topic wave functions $\left|\sigma_{k}\right\rangle $
represented in the term space, and the other one representing the
interaction between documents and latent topics. 

Using \ref{eq:term-doc-prob} and \ref{eq:wave-function-cod-QLSA}
we can calculate the approximation of $P(t_{j}|d_{i})$ generated
by QLSA:
\begin{eqnarray}
P(t_{j}|d_{i}) & \approx & \left[\sum_{i=1}^{r}\left\langle \tau_{j}|\sigma_{k}\right\rangle \frac{\left\langle \sigma_{k}|\varphi_{i}\right\rangle }{\left\Vert P_{S}\left|\varphi_{i}\right\rangle \right\Vert }\right]^{2}\nonumber \\
 & = & \sum_{i=1}^{r}\left\langle \tau_{j}|\sigma_{k}\right\rangle ^{2}\frac{\left\langle \sigma_{k}|\varphi_{i}\right\rangle ^{2}}{\left\Vert P_{S}\left|\varphi_{i}\right\rangle \right\Vert ^{2}}+I_{ji}\nonumber \\
 & = & \sum_{k=1}^{r}P(t_{j}|z_{k})P(z_{k}|d_{i})+I_{ji},\label{eq:QLSA-decomposition}
\end{eqnarray}
 where $I_{ji}=\left[\sum_{k,l=1\dots r,k\ne l}\left\langle \tau_{j}|\sigma_{k}\right\rangle \left\langle \sigma_{k}|\varphi_{i}\right\rangle \left\langle \tau_{j}|\sigma_{l}\right\rangle \left\langle \sigma_{l}|\varphi_{i}\right\rangle \right]/\left\Vert P_{S}\left|\varphi_{i}\right\rangle \right\Vert ^{2}$.
Checking \ref{eq:PLSA-decomposition} and \ref{eq:QLSA-decomposition}
it is easy to see the difference between both approximations, QLSA
adds the additional term $I_{ji}$. This term could be interpreted
as an interference term \cite{Zuccon2010}.

\section{Experimental Evaluation}

In this section we perform an exploratory experimentation that evaluates
the performance of QLSA against LSA. As discussed in Section \ref{subsec:QLSA-vs-LSA},
both methods share a common geometrical approach that finds a low-dimensional
space using SVD. The main difference resides in the document representation
used. Thus, the goal of the experimental evaluation is to establish
the effect of the quantum representation when using a latent topic
indexing strategy for document retrieval.

In our experiments we evaluated the automatic indexing task to support
query based retrieval. The performance is measured in terms of Mean
Average Precision (MAP) for two standard datasets to assess the empirical
differences between the formulated method and two baseline approaches:
direct matching in a Vector Space Model, using cosine similarity,
and the LSA approach. The experimental setup is intentionally kept
simple, only term frequency is used without any kind of weighting,
simple stop-word removal and stemming preprocessing is applied. Document
search is performed by projecting the query terms and using the cosine
similarity with respect to other documents in the latent space, i.e.,
ranking scores are taken directly from the latent space. 

\subsection{Collections}

To follow an evaluation of ranked retrieval, we used three collections
with relevance assessment: (1) the MED collection, a common dataset
used in early information retrieval evaluation, composed of 1033 medical
abstracts and 30 queries, all indexed with about 7000 terms; (2) the
CRAN collection, another standard dataset with 1400 document abstracts
on aeronautics from the Cranfield institute of Technology and 225
queries, is indexed with about 3700 terms. (3) The CACM collection,
with 3204 abstracts from the Communications of the ACM Journal with
64 queries, is indexed with about 3000 terms.

\subsection{Dimensions of the Latent Space}

\begin{figure}
\begin{centering}
\subfloat[\label{fig:MED-factors}MED]{\begin{centering}
\includegraphics[width=0.5\textwidth]{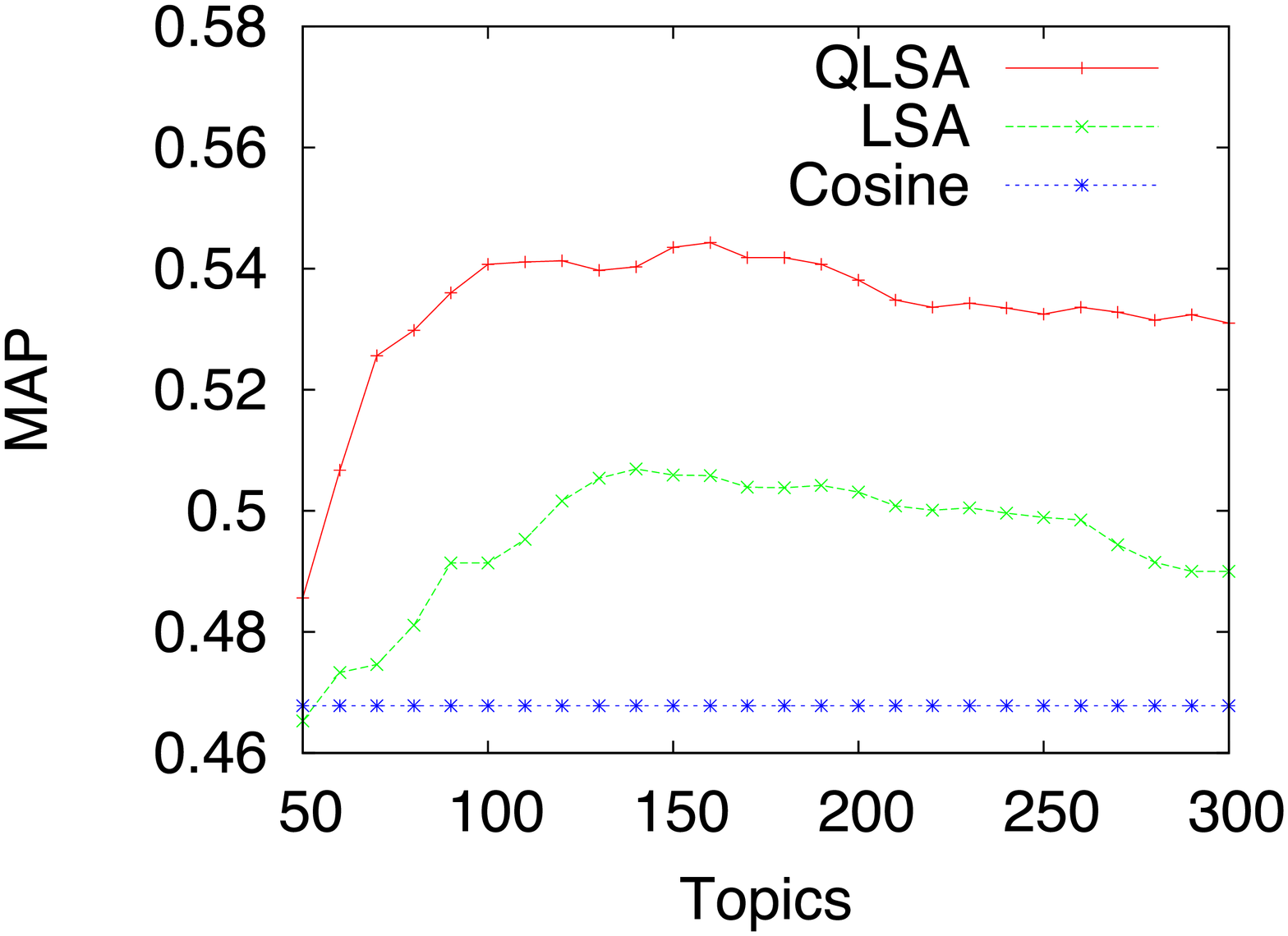}
\par\end{centering}
}
\par\end{centering}
\begin{centering}
\subfloat[\label{fig:CRAN-factors}CRAN]{\begin{centering}
\includegraphics[width=0.5\textwidth]{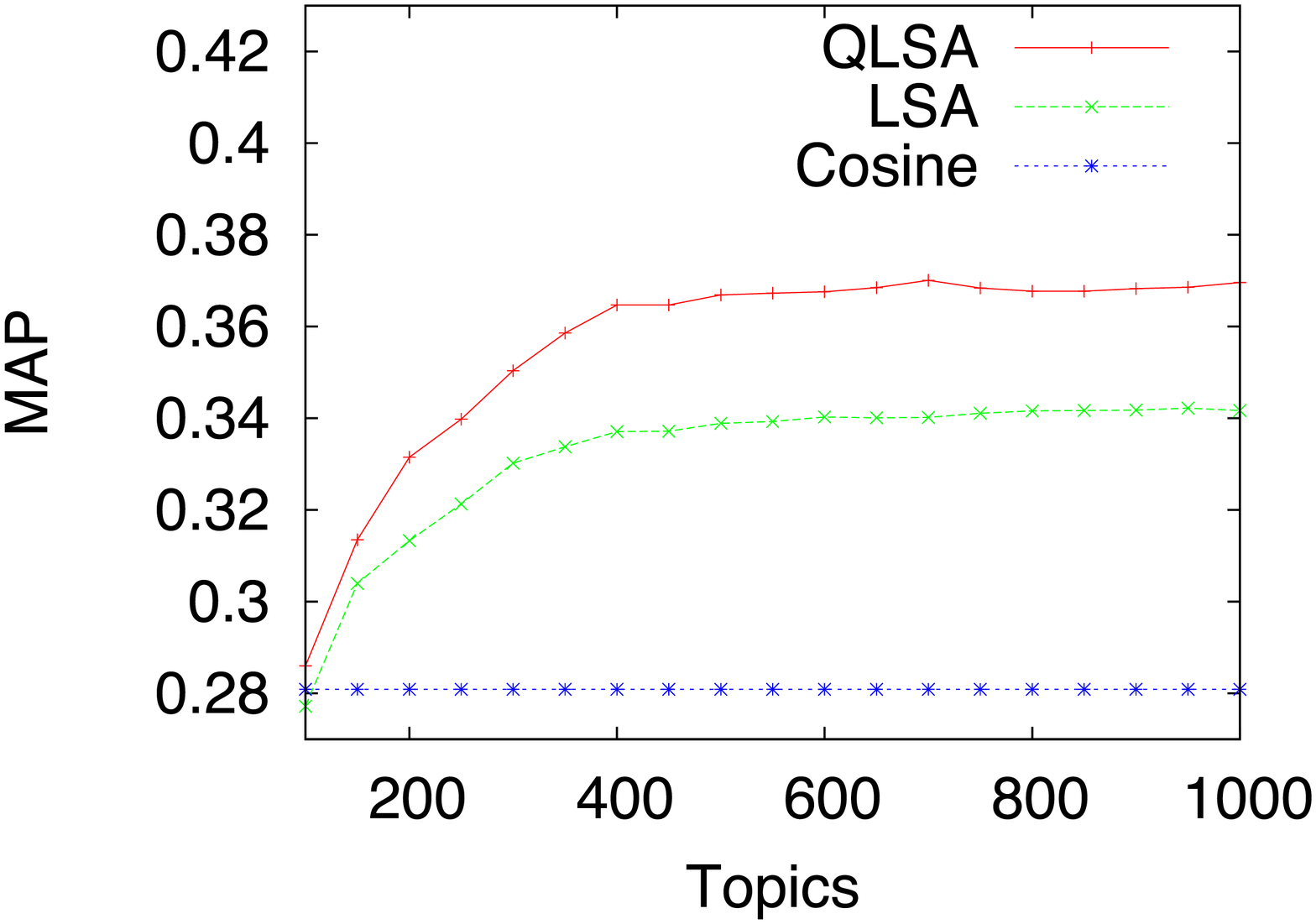}
\par\end{centering}
}\subfloat[\label{fig:CACM-factors}CACM]{\begin{centering}
\includegraphics[width=0.5\textwidth]{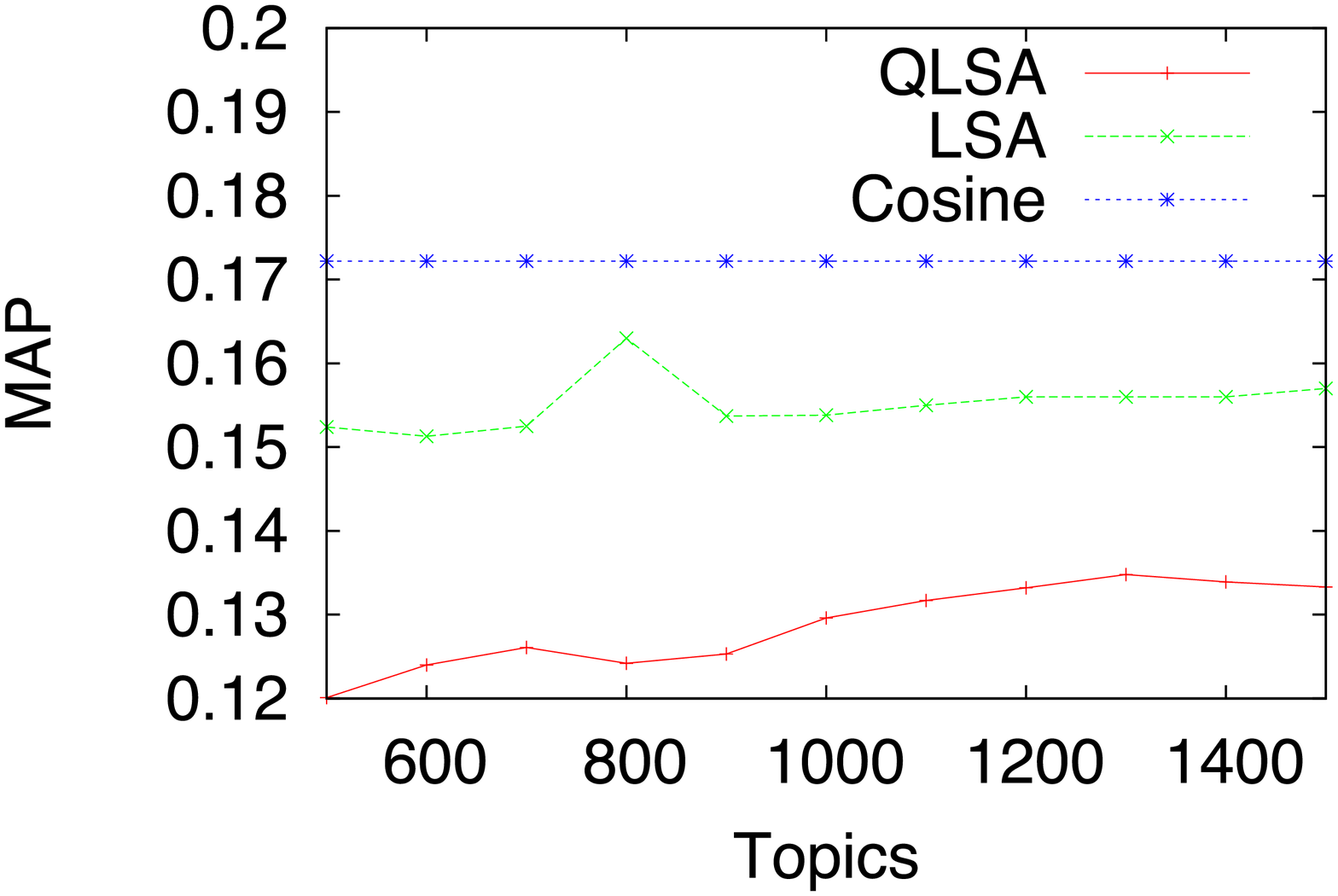}
\par\end{centering}
}
\par\end{centering}
\centering{}\caption{\label{fig:all-factors}Variation of number of topics for the different
collections}
\end{figure}

Figure \ref{fig:all-factors} presents the variation of MAP with respect
to the number of latent factors for the evaluated collections. It
shows that latent indexing methods provide an improvement over the
cosine similarity baseline for the MED and CRAN collections. The dimension
of the latent space was varied from 50 to 300 factors taking steps
of 10 units for the MED collection and from 100 to 1000 factors taking
steps of 50 units for the CRAN collection. The CACM collection, however,
does not show improvements when using latent factors for document
indexing.

For the first two collections, results show that QLSA performs better
than LSA for every evaluated dimension of the latent topic space.
In the MED collection, the performance of both methods increases to
reach a maximum value around the same latent space dimensionality
(between 140 and 160) and then starts to decrease slowly again. In
the CRAN collection, the performance of both methods increases and
tends to get stable after 500 topics. The best number of topics is
very similar for both methods, however, the performance is significantly
improved in favor of QLSA.

The CACM collection is particularly challenging for LSA, and QLSA
does not perform better. In fact, QLSA seems to amplify the bad performance
of LSA. In the case of LSA, this is consistent with previously reported
performances in the literature, that showed no benefit for query based
retrieval, but instead, a decreasing in performance. 

\subsection{Recall-Precision Evaluation}

\begin{figure}[p]
\begin{centering}
\subfloat[MED]{\begin{centering}
\includegraphics[width=0.5\textwidth]{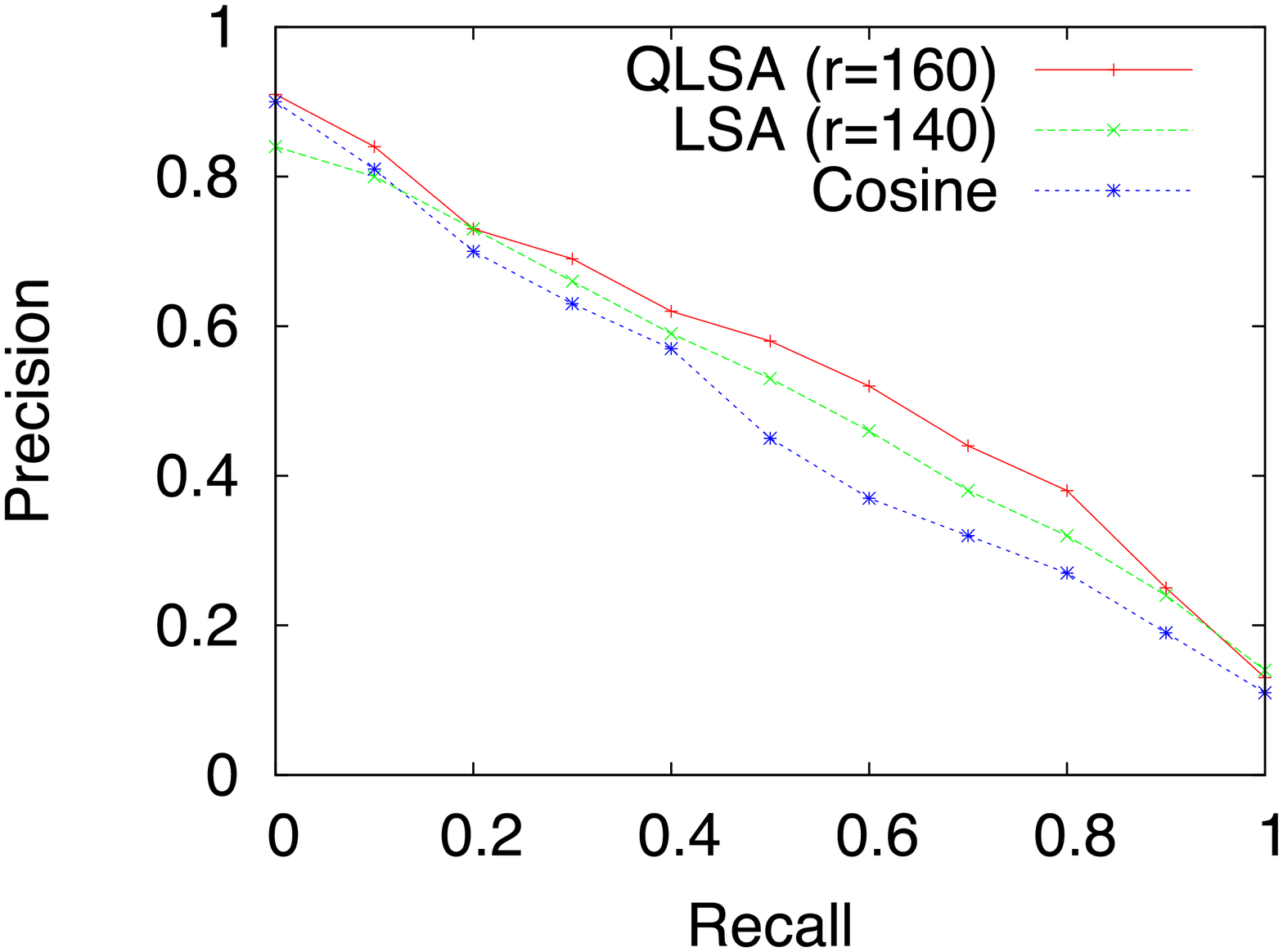}
\par\end{centering}
}
\par\end{centering}
\begin{centering}
\subfloat[CRAN]{\begin{centering}
\includegraphics[width=0.5\textwidth]{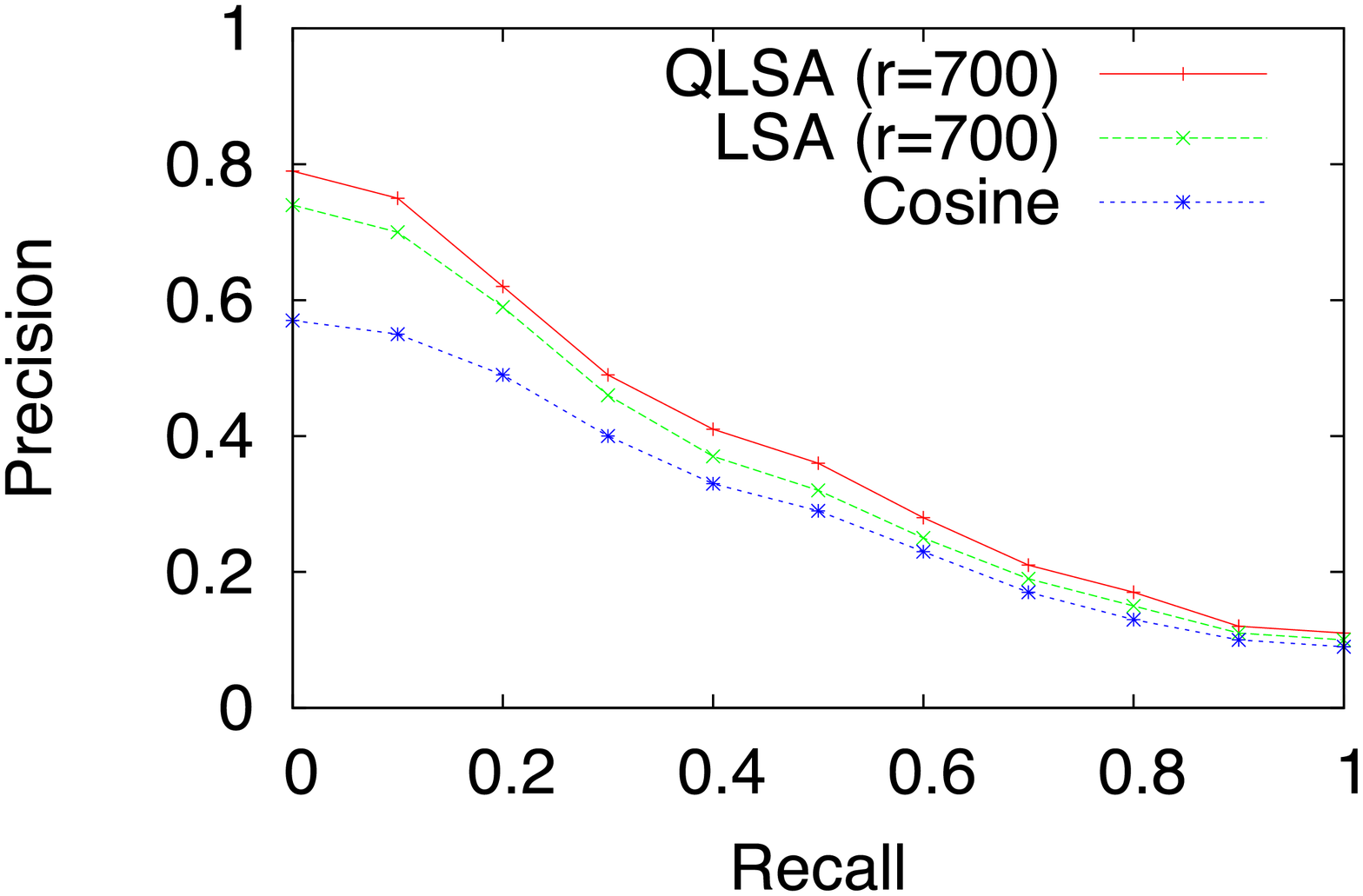}
\par\end{centering}
}\subfloat[CACM]{\begin{centering}
\includegraphics[width=0.5\textwidth]{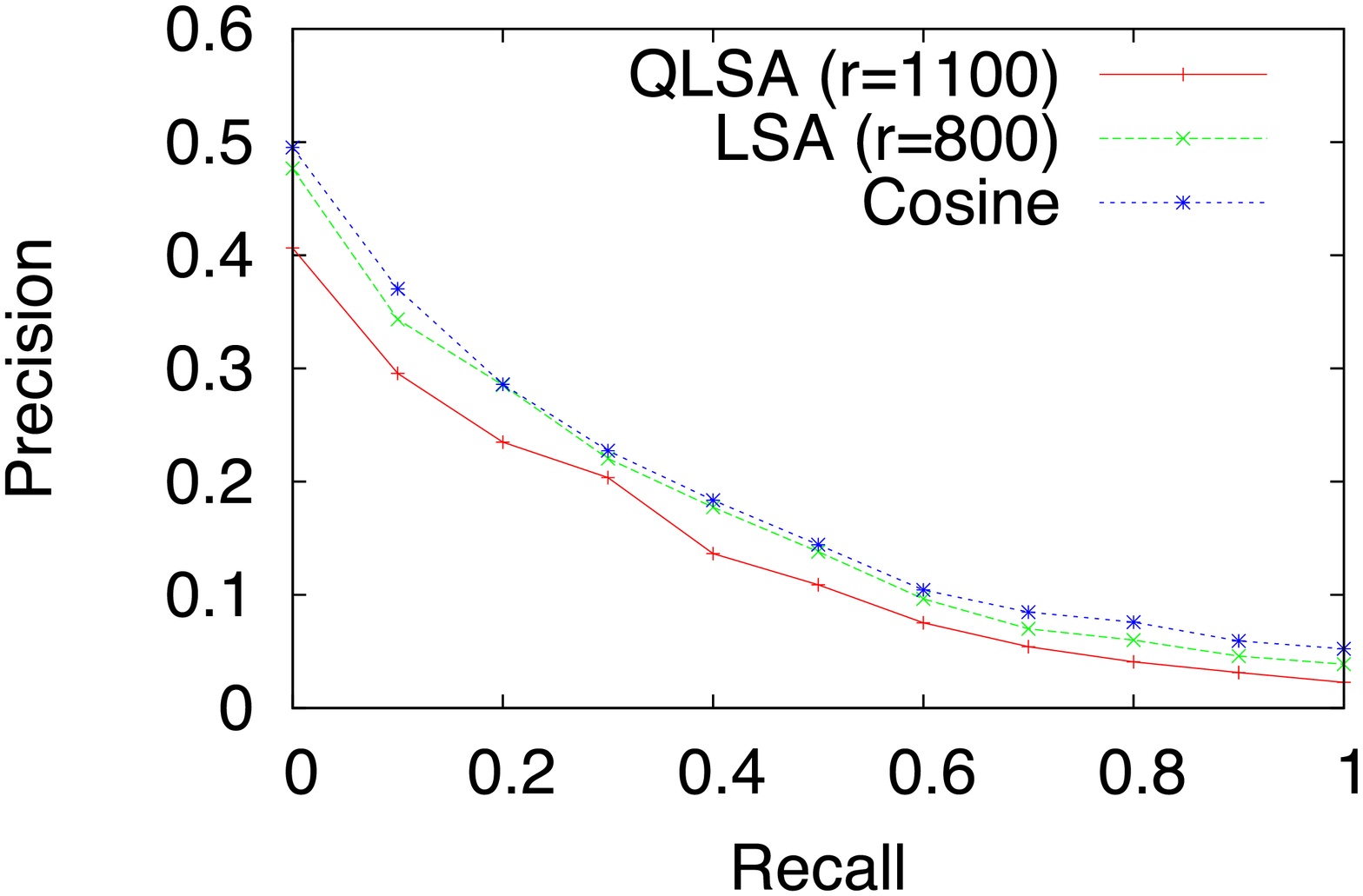}
\par\end{centering}
}
\par\end{centering}
\caption{\label{fig:all-Recall-Precision}Recall-Precision graphs for the three
collections and three methods with the best latent factor dimensions
in each case.}
\end{figure}

Figure \ref{fig:all-Recall-Precision} shows the interpolated Recall-Precision
graph for the 3 evaluated approaches, averaged over the available
set of queries. Each model has been configured with the best latent
space dimensionality, according to the analysis on the previous Section.
Again, results show that latent topic indexing provides a better response
over the direct matching approach in the MED and CRAN collections.
The plots also show an improved response of QLSA over both cosine
and LSA approaches, in these two collections.

In the MED collection, QLSA provides a slightly better response with
respect to the cosine similarity in the early stages of the retrieval
process, and then starts to show a larger improvement. LSA starts
worse than cosine but after the first part of the results it overtakes
the baseline and shows a better response in the long term retrieval.
QLSA presents a better response than LSA during the whole retrieval
process. In the case of the CRAN collection, QLSA and LSA show a general
improvement over the baseline, both in the early and long term retrieval.
QLSA again offers better results than the other two methods, showing
a consistent improvement in terms of precision for the ranked retrieval
task.

Figure \ref{fig:all-Recall-Precision}-c shows the response of the
indexing methods on the CACM collection, showing an important decreasing
for QLSA. We hypothesize that, for this collection, discriminative
terms are mixed with other terms in latent factors, leading to a lose
of discerning capacity of the ranking method.

\begin{table}[p]
\caption{\label{tab:Summary-map}Summary of the retrieval performance on the
test collections. Reported values are Mean Average Precision over
all the available queries.}

\centering{}%
\begin{tabular}{|c|c|c|c|c|c|c|}
\cline{2-7} 
\multicolumn{1}{c|}{} & \multicolumn{2}{c|}{MED} & \multicolumn{2}{c|}{CRAN} & \multicolumn{2}{c|}{\emph{CACM}}\tabularnewline
\hline 
\emph{Method} & \emph{Precision} & \emph{Improv.} & \emph{Precision} & \emph{Improv.} & \emph{Precision} & \emph{Improv.}\tabularnewline
\hline 
cosine & 0.4678 & - & 0.2809 & - & 0.1722 & \emph{-}\tabularnewline
\hline 
LSA & 0.5069 & +8.36\% & 0.3302 & +17.55\% & 0.1630 & -5.34\%\tabularnewline
\hline 
QLSA & 0.5443 & +16.35\% & 0.3504 & +24.74\% & 0.1315 & -23.64\%\tabularnewline
\hline 
\end{tabular}
\end{table}

Table \ref{tab:Summary-map} summarizes the results obtained in this
exploratory evaluation, showing that QLSA results in an important
improvement with respect to LSA for two collections even though both
algorithms are based on a SVD. These results complement the theoretical
differences between both algorithms and highlight the empirical benefits
of using a QIR-based algorithm for modelling latent topics. In the
case of the CACM collection, both LSA and QLSA show a decreasing in
performance with respect to the baseline, with a larger margin for
QLSA. It is interesting to see that when LSA performs better than
the baseline, QLSA is able to outperform both, the baseline and LSA.
But, when LSA does not improve, QLSA performs even worse.

A comparison against PLSA was not performed, however, the results
reported by \cite{Hofmann1999} could serve as a reference, despite
they were obtained with a slightly different experimental setup that
favors the performance of the algorithms. It reports an average precision
of 63.9, 35.1 and 22.9 for MED, CRAN and CACM respectively, using
PLSA. According to these results, QLSA does not outperforms PLSA,
however, it shows a competitive performance on two of the datasets,
on the other one the performance was remarkable bad. 

\section{Discussion and Conclusions}

Given its exploratory nature, the experimental results are not conclusive.
However, the results are encouraging and suggest that the quantum
representation could provide a good foundation for latent topic analysis.
The approaches followed by both QLSA and LSA are very similar, the
main difference is the document representation used. It is interesting
to see the effect of the quantum representation on LSA performance:
it improved the performance on two of the datasets where LSA showed
some advantage over the baseline, but also it amplified the bad performance
on the other dataset. However, QLSA has a clear advantage over LSA,
its more principled representation of the geometry of the document
space allows a probabilistic interpretation.

LTA methods based on probabilistic modeling, such as PLSA and LDA,
have shown better performance than geometry-based methods. However,
with methods such as QLSA it is possible to bring the geometrical
and the probabilistic approaches together. Here we started from a
geometrical stand point to formulate the model and then we provided
a probabilistic interpretation of it. Thanks to the dual nature of
the quantum representation, it is possible to do exactly the opposite:
start from a probabilistic latent topic model and then give it a geometrical
interpretation. A good start point would be the theory of quantum
probabilistic networks \cite{Tucci1997,LaMura2007,Laskey2007}.

There are many remaining open questions that justify further investigation:
what is the interpretation of the interference term (Eq. \ref{eq:QLSA-decomposition})
in the approximation of $P(t_{j}|d_{i})$ generated by QLSA? How to
implement quantum versions of probabilistic LTA methods such as PLSA
and LDA? These questions are the main focus of our ongoing research
work.

\bibliographystyle{plain}
\bibliography{qlsa}

\end{document}